\documentclass{article}

    \PassOptionsToPackage{numbers, compress}{natbib}

    \usepackage[preprint]{neurips_2023}

\usepackage[utf8]{inputenc} 
\usepackage[T1]{fontenc}    
\usepackage[pagebackref,breaklinks,colorlinks]{hyperref}
\usepackage{url}            
\usepackage{booktabs}       
\usepackage{amsfonts}       
\usepackage{nicefrac}       
\usepackage{microtype}      
\usepackage{xcolor}
\usepackage{makecell}
\usepackage{array}
\usepackage{color}
\usepackage{colortbl}
\usepackage{wrapfig}

\usepackage{amsmath}
\usepackage{amssymb}
\usepackage{multirow}
\usepackage{caption}
\usepackage{subfloat}
\usepackage{graphicx}
\usepackage[caption=false, font=footnotesize]{subfig}
\usepackage{filecontents}
\usepackage{pgfplots}
\usepackage{tikzscale}
\usepackage{tikz}
\usepgfplotslibrary{patchplots}
\usepackage[capitalize]{cleveref}
\Crefname{section}{Sec.}{Secs.}
\Crefname{section}{Section}{Sections}
\Crefname{table}{Table}{Tables}
\Crefname{table}{Tab.}{Tabs.}

\definecolor{white}{RGB}{.255,.255,.255}
\definecolor{black}{RGB}{.84, .255, .159}

\title{SOC: Semantic-Assisted  Object Cluster for \\Referring Video Object Segmentation}

\author{%
  Zhuoyan Luo$^{1}$\thanks{Equal contribution.
  $\dagger$ Corresponding author.} , \qquad 
   Yicheng Xiao$^{1\ast}$, \qquad
  Yong Liu$^{12\ast}$, \qquad
  \\
  \textbf{Shuyan Li$^{3}$},
  \textbf{Yitong Wang$^{2}$},
  \textbf{Yansong Tang$^{1\dagger}$},
  \textbf{Xiu Li$^{1\dagger}$},
  \textbf{Yujiu Yang$^{1\dagger}$} \\
  \\$^{1}$Tsinghua Shenzhen International Graduate School, Tsinghua University
  \\$^{2}$ByteDance Inc.\\
  $^{3}$Engineering Department, University of Cambridge \\
  \texttt{\{robertluo171, yichengxiao888\}@gmail.com, liu-yong20@mails.tsinghua.edu.cn}
}

\begin{document}

\maketitle

\begin{abstract}
  This paper studies referring video object segmentation (RVOS) by boosting video-level visual-linguistic alignment. 
  Recent approaches model the RVOS task as a sequence prediction problem and perform multi-modal interaction as well as segmentation for each frame separately.  
  However, the lack of a global view of video content leads to difficulties in effectively utilizing inter-frame relationships and understanding textual descriptions of object temporal variations.
  To address this issue, we propose \textbf{S}emantic-assisted \textbf{O}bject \textbf{C}luster (SOC), which aggregates video content and textual guidance for unified temporal modeling and cross-modal alignment.
  By associating a group of frame-level object embeddings with language tokens, SOC facilitates joint space learning across modalities and time steps.
  Moreover, we present multi-modal contrastive supervision to help construct well-aligned joint space at the video level. 
  We conduct extensive experiments on popular RVOS benchmarks, and our method outperforms state-of-the-art competitors on all benchmarks by a remarkable margin. 
  Besides, the emphasis on temporal coherence enhances the segmentation stability and adaptability of our method in processing text expressions with temporal variations. Code will be available.
\end{abstract}

\section{Introduction}
Referring Video Object Segmentation (RVOS)~\cite{mttr,referformer} aims to segment the target object referred by the given text description in a video. 
Unlike conventional single-modal segmentation tasks according to pre-defined categories~\cite{vipseg, vspw} or visual guidance~\cite{vos2,vos1}, referring segmentation requires comprehensive understanding of the content across different modalities to identify and segment the target object accurately. Compared to referring image segmentation, RVOS is even more challenging since the algorithms must also model the temporal relationships of different objects and locations. This emerging topic has attracted great attention and has many potential applications, such as video editing and human-robot interaction.

\begin{figure*}
    \centering
    \includegraphics[width=\linewidth]{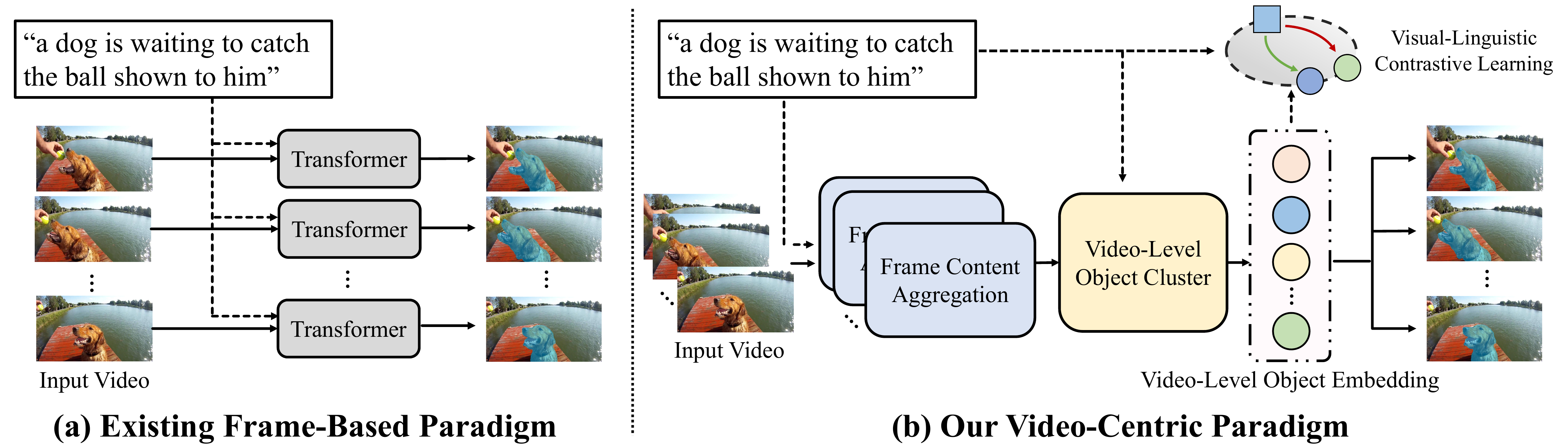}
    \caption{Illustration of different paradigms. Frame-based methods perform cross-modal interaction and segmentation for each frame individually. In contrast, our method unifies temporal modeling and cross-modal alignment to achieve video-level understanding.}
    \label{fig:teaser}
    \vspace{-10pt}
\end{figure*}

Due to the varieties of video content as well as the unrestricted language expression, the critical problem of RVOS lies in how to perform pixel-level alignment between different modalities and time steps.
To accomplish this challenging task, previous methods have tried various alignment workflows. Early approaches~\cite{vlt,GKC,bottom2,top1,bottom3,simbaseline,bottom4} take the bottom-up or top-down paradigms to segment each frame separately, while recent works~\cite{mttr, referformer} propose to unify cross-modal interaction with pixel-level understanding into transformer structure. 
Although the above-mentioned approaches have facilitated the alignment between different modalities and achieved excellent performance, they model the RVOS task as a sequence prediction problem and pay little attention to the temporal relationships between different frames. Specifically, they perform cross-modal interaction and segmentation for each frame individually, as illustrated in \cref{fig:teaser} (a). The exploitation of temporal guidance relies on the spatial-temporal backbone and manually designed hard assignment strategies~\cite{mttr,referformer}.
Such paradigms convert the referring video segmentation into stacks of referring image segmentation. 
While it may be acceptable for such conversion to handle the descriptions of static properties such as the appearance and color of the objects, this approach may lose the perception of target objects  for language descriptions expressing temporal variations of objects due to the lack of video-level multi-modal understanding.

To alleviate the above problems and align video with text effectively, we propose Semantic-assisted Object Cluster (SOC) to perform object aggregation and promote  visual-linguistic alignment at the video level, as depicted in \cref{fig:teaser} (b).
Specifically, we design a Semantic Integration Module (SIM) to efficiently aggregate intra-frame and inter-frame information. With a global view of the video content, SIM can facilitate the understanding of temporal variations as well as alignment across different modalities and granularity.
Furthermore, we introduce visual-linguistic contrastive learning to provide semantic supervision and guide the establishment of video-level multi-modal joint space. 
In addition to the remarkable improvements in generic scenarios, these efforts also allow our method to effectively handle text descriptions expressing temporal variations.

We conduct experiments on popular RVOS benchmarks, \textit{i.e.}, Ref-YouTube-VOS~\cite{urvos}, Ref-DAVIS~\cite{davis}, A2D-Sentences and JHMDB-Sentences~\cite{GavrilyukGLS18}, to validate the effectiveness of our method.
Results show that SOC notably outperforms existing methods for all benchmarks with  faster inference speed.
In addition, we provide detailed ablations and analysis on components of our method.

Overall, our contributions are summarized as follows:
\begin{itemize}
    \item We present a framework called SOC for RVOS to unify temporal modeling and cross-modal alignment. In SOC, a Semantic Integration Module (SIM) is designed to efficiently aggregate inter and intra-frame information, which achieves video-level multi-modal understanding. 
    \item We introduce a visual-linguistic contrastive loss to apply semantic supervision on video-level object representations, resulting in well-aligned multi-modal joint space. 
    \item Without bells and whistles, our method outperforms existing state-of-the-art method ReferFormer~\cite{referformer} by a remarkable margin, \textit{e.g.}, +3.0\% $\mathcal{J} \& \mathcal{F}$ on Ref-YouTube-VOS and +3.8\% $\mathcal{J} \& \mathcal{F}$ on Ref-DAVIS under fair comparison. Besides, our method runs at 32.3 FPS on single 3090 GPU, which is significantly faster than the 21.4 FPS of ReferFormer.
\end{itemize}

\section{Related Work}
\paragraph{Referring Image Segmentation}
Referring Image Segmentation (RIS) aims to localize the corresponding object referred by a text description within a static image. It is first introduced by Hu \textit{et al.}~\cite{hu16}, who developes a simple framework that utilizes Fully Convolution Network (FCN)~\cite{fcn} to generate segmentation masks from concatenated visual and linguistic features. To deeply explore the intrinsic correlations among different modal features, several studies~\cite{Ding19, hu20, cris, yecmsa} design various attention modules for modality interaction. Additionally, VLT~\cite{vlticcv} proposes a transformer-based architecture for the RIS task, which has gained more popularity than FCN-based approaches. LAVT~\cite{lavt} incorporates early alignment of visual and linguistic features at the intermediate layers of encoders. PolyFormer~\cite{polyformer} further uses transformer to generate polygon vertices as the prior information to refine segmentation masks, leading to better results.      

\paragraph{Referring Video Object Segmentation}
Compared to RIS, RVOS is more challenging since both the action and appearance of the referred object are required to be segmented in a dynamic video. 
Gavrilyuk \textit{et al.} ~\cite{GavrilyukGLS18} first proposes the Referring Video Object Segmentation (RVOS) task. URVOS~\cite{urvos} introduces a large-scale RVOS benchmark and a unified framework that leverages attention mechanisms and mask propagation to increase the task's complexity and scope. ACAN~\cite{wang2019acan} designs an asymmetric cross-guided attention network to establish complex visual-linguistic relationships.
To improve positional relation representations in the text, PRPE~\cite{prpe} explores a positional encoding mechanism based on the polar coordinate system.
In addition, most previous approaches~\cite{liang2021clawcranenet, liu2022cmpc, rvos1, cmfsa, rvos2, vtcapsule} rely on complicated pipelines. To simplify the workflow, MTTR~\cite{mttr} and ReferFormer~\cite{referformer} adopt query-based end-to-end frameworks for decoding objects from multi-modal features, achieving excellent performance. However, during the decoding phase, previous methods only concentrate on intra-frame object information, disregarding the valuable temporal context of objects across frames. To address this issue, we propose to associate object temporal context with language tokens and achieve video-level multi-modal understanding. 

\begin{figure*}[t]
    \centering
    \includegraphics[width=\linewidth]{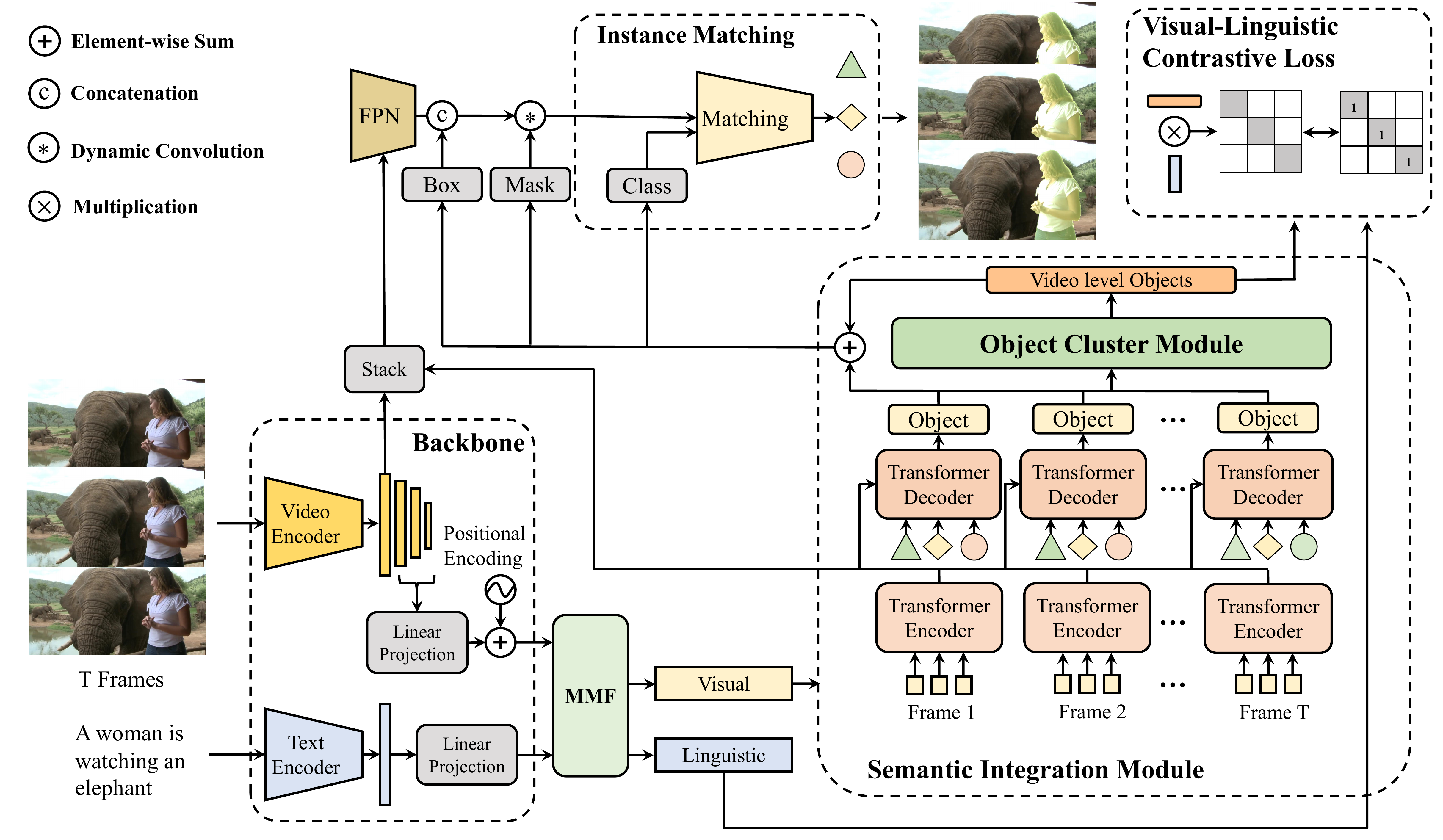}
    \vspace{-15pt}
    \caption{Overview of SOC. The model takes a video clip with corresponding language descriptions as input. After the encoding process, the multi-modal fusion (MMF) module performs bidirectional fusion to build the intrinsic feature relations. Then we design a Semantic Integration Module to efficiently aggregate intra-frame and inter-frame information. Meanwhile, we introduce a visual-linguistic contrastive loss to benefit the establishment of video-level multi-modal space. Finally, the prediction heads decode the condensed embeddings and output segmentation masks.}
    \label{fig:framework}
    \vspace{-5pt}
\end{figure*}

\section{Method}

Given $T$ frames of video clip $\mathcal{I} = \{I_t\}_{t=1}^T$, where $I_t \in \mathbb{R}^{3 \times H_0 \times W_0}$ and a referring text expression $\mathcal{E} = \{e_i\}_{i=1}^L$, where $e_i$ denotes the i-th word in the text. 
Our goal is to generate a series of binary segmentation masks $\mathcal{S} = \{s_t\}_{t=1}^T$, $s_t \in \mathbb{R}^{1\times H_0 \times W_0}$ of the referred object.
To this end, we propose a video-centric framework called Semantic-assisted Object Cluster (SOC). We will elaborate on it in the following sections.

\subsection{Visual and Linguistic Encoding}
\paragraph{Visual Encoder}
Taking a video clip $\mathcal{I}$ as input, we utilize a spatial-temporal backbone such as Video Swin Transformer~\cite{videoswin} to extract hierarchical vision features.
Consequently, the video clip is encoded into a set of feature maps $\mathcal{F}^v_i\in \mathrm{R}^{C_i \times H_i \times W_i} $, $i \in \{1, 2, 3, 4\}$. Here $H_i$ and $W_i$ denote the height and width of each scale feature map, respectively. $C$ denotes the channel dimension.

\vspace{-5pt}
\paragraph{Language Encoder}
Simultaneously, a transformer-based~\cite{transformer} language encoder encodes the given textual expression $\mathcal{E}$ to a word-level embedding $\mathcal{F}^w \in \mathbb{R}^{L\times C_t}$ and a sentence-level embedding $\mathcal{F}^s\in \mathbb{R}^{1\times C_t}$.
The word embedding $\mathcal{F}^w$ contains fine-grained description information.
On the other hand, the sentence embedding $\mathcal{F}^s$ expresses the general characteristics of the referred target object.

\subsection{Two Stream Multi-Modal Fusion}
Having the separate visual and linguistic embedding encoded from the video clip and text expression, we design a Multi-Modal Fusion module called MMF to perform preliminary cross-modal alignment.
As shown in \cref{fig:module}, MMF is a two-stream structure.
The language-to-vision (L2V) stream aims to highlight the corresponding regions of the referred object in each frame and mitigate the effect of background noise.
It leverages linguistic information as guidance and addresses the potential similar visual areas. 
Meanwhile, a vision-to-language (V2L) stream is designed to update the textual embedding with image content, which helps to relieve the potential ambiguity of unconstrained descriptions. 
Specifically, we measure the relevance of all visual areas to the text query and assign weights to the useful information extracted from the visual features so as to reorganize the text embedding.
The above L2V and V2L fusion process are based on the multi-head cross-attention mechanism, which can be formulated as:
\begin{equation}
    \begin{gathered}
    \mathrm{MHA}\left(\mathcal{X}, \mathcal{Y}\right) = \mathrm{Concat}\left(head_1\left(\mathcal{X},\mathcal{Y}\right),\dots,head_h\left(\mathcal{X},\mathcal{Y}\right)\right)W, \\ 
    head_j\left(\mathcal{X}, \mathcal{Y}\right) = \mathrm{softmax}\left(\frac{\left(\mathcal{X}W^Q_j\right)^T\mathcal{Y}W^K_j}{\sqrt{C}}\right) \mathcal{Y}W^V_j,
    \end{gathered}
\end{equation}
where $\mathrm{MHA}\left(\cdot \right)$ stands for multi-head attention. 
$W, W^Q_j, W^K_j, W^V_j$ are learnable weights used to map the input to the attention space.

Since visual features of different scales contain diverse content information, MMF is designed to produce a series of coarse-to-fine visual and textual feature maps $\{\mathcal{F}^{vf}_i\}$ and $\{\mathcal{F}^{ef}_i\}$, $i \in \{2, 3, 4\}$. Specifically, we leverage shared parameters to perform multi-head cross-attention operations on $\{\mathcal{F}^v_i\}$, $i \in \{2, 3, 4\}$ and $\mathcal{F}^w$. 
Take $\mathcal{F}^v_{2}$ as an example. 
Firstly, we utilize $1\times 1$ convolution and fully connected layers to transform the visual and linguistic embeddings into joint space, respectively. Then the bidirectional multi-modal fusion is applied to align information from different modalities as well as enhance the feature representation. The fusion process is:
\begin{gather}
    \mathcal{F}^{vf}_{2} = \mathrm{MHA}\left(\mathcal{F}^{v}_{2}, \mathcal{F}^w\right) \cdot \mathcal{F}^{v}_{2}, \label{eq:2}  \\ 
    \mathcal{F}^{ef}_{2} = \mathrm{MHA}\left(\mathcal{F}^{w}, \mathcal{F}^{v}_{2} \right) \cdot \mathcal{F}^{w}, \label{eq:3}
\end{gather}
where $\mathcal{F}^{v}_{2} \in \mathbb{R}^{TH_2W_2 \times D}$ and  $\mathcal{F}^w \in \mathbb{R}^{L\times D}$ are embeddings projected by convolution and fully connected layers.
Similar to \cref{eq:2} and \cref{eq:3}, the coarse-to-fine aligned visual features $\{\mathcal{F}^{vf}_i\}$ and textual embeddings $\{\mathcal{F}^{ef}_i\}$, $i \in \{2,3,4\}$ are produced.

\subsection{Semantic Integration Module
}
After aligning cross-modal information and activating the potential target region in MMF, we design a Semantic Integration Module (SIM) to incorporate visual content and generate compact target representations. Specifically, we first leverage frame-level content aggregation to locate objects separately for each frame. Then we associate inter-frame dependency and model the temporal relationship of objects via video-level object cluster.

\vspace{-5pt}
\paragraph{Frame-Level Content Aggregation}
Having the activated visual features $\{\mathcal{F}^{vf}_i\}$ from MMF, we utilize a transformer-based structure to locate objects in each frame. 
Firstly, $K$ stacks of deformable transformer encoder layer~\cite{deformabledetr} are leveraged to capture intra-frame relationships and further excavate multi-modal interactions inside $\{\mathcal{F}^{vf}_i\}$.
This process can be formulated as:
\begin{equation}
\mathcal{F'}^{vf}_i = \left\{\mathrm{DecformEnc}_{k}\left(f^{vf}_t\right) \right\}_{t=1}^T,
\label{eq:4}
\end{equation}
where $f^{vf}_t$ denotes the activated visual features of the t-th frame.
Then, a set of learnable object queries~\cite{detr, deformabledetr} is introduced to aggregate image content and highlight potential target objects. 
Following~\cite{referformer}, these object queries fully interact with $\mathcal{F'}^{vf}_i$ in deformable transformer decoder through cross-attention mechanism.
After extracting different object representations, these object queries are turned into instance embeddings $\mathcal{O}^{f} \in \mathbb{R}^{T \times N_q \times D}$.
Note that we set up $N_q$ object queries to represent instances of each frame in a video clip so there are $N_qT$ output object queries in total.

\begin{figure*}
    \centering
    \includegraphics[width=\linewidth]{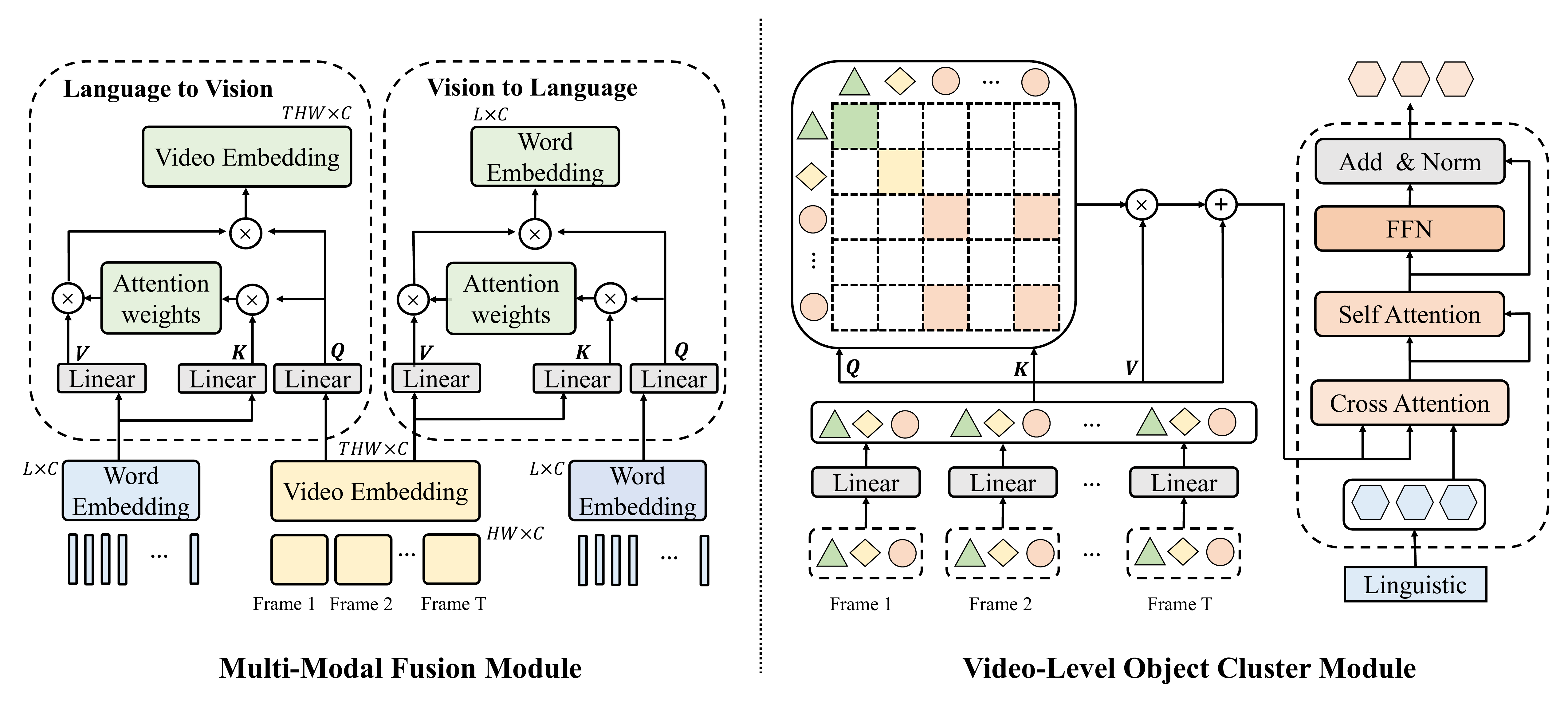}
    \vspace{-10pt}
    \caption{The structure of proposed  Multi-Modal Fusion module (MMF) and Video-level Object Cluster module (VOC).}
    \label{fig:module}
    \vspace{-10pt}
\end{figure*}

\vspace{-5pt}
\paragraph{Video-Level Object Cluster}\label{par:voc}
Instances typically vary in pose and location between frames and even being obscured. Previous methods~\cite{mttr, vlt, lbdt, referformer} only model instances separately for each frame, disregarding the temporal relationship and continuous motion of instances. Although such an approach can cover simple scenarios such as descriptions of target appearance, the lack of inter-frame interaction makes existing methods inefficient for the description of temporal relationships.
To address the aforementioned weakness, inspired by ~\cite{sstvos, vita, videoknet}, we design a video-level object cluster to capture the temporal information of instances between frames.

As shown in \cref{fig:module}, after instances embedding $\mathcal{O}^{f}$ are formulated by the frame-level content aggregation, we flatten it into $\mathcal{O}^{f'} \in \mathbb{R}^{TN_q \times D}$ and employ self-attention mechanism along the temporal axis to introduce inter-frame interaction.
Furthermore, we find that only introducing temporal self-attention is inferior.
Simply sharing temporal object context may create redundancy due to the similarity of the representation referring to the same object in different frames, which potentially affects the model's precise understanding of objects in the video clip.
To this end, we employ an object grouping decoder to perform object clustering and group the same object into video-level compact representations across frames.
Specifically, we introduce $N_v$ video-level object queries $\mathcal{O}^{v} \in \mathbb{R}^{N_v \times D}$ and they are initialized by linguistic sentence-level feature $\mathcal{F}^s$, which helps to promote the understanding of object descriptions and build the visual-linguistic joint space.
By performing interaction between linguistic-aware video queries and condensed instance information of each frame, the object grouping decoder can effectively capture temporal object contexts and group the referred object queries across frames to output the clustered video-level object queries $\mathcal{O}^{v}$. 
While temporal connections can provide a lot of valuable information, the segmentation of each frame is dependent on the specific image content.
Thus, we incorporate the high-level video instance information into each frame to integrate the advantages of both.
In detail, as illustrated in \cref{fig:framework}, we repeat the video-level object queries $T$ times $\mathcal{O'}^{v} \in \mathbb{R}^{T\times N_v \times D}$ and enhance the representation of frame-level object queries by element-wise sum: $\mathcal{O}^{f} = \mathcal{O}^{f} + \mathcal{O'}^{v}$. In this way, the semantic of frame-level object queries is greatly enriched by the supplement of video-level object information.

\subsection{Visual-Linguistic Contrastive Learning}\label{sec:3.4}
Although the aggregated video-level representation can better describe object states, the simple exploitation of textual priors may lead to non-target  potential responses for video-level embeddings. Meanwhile, taking unrestricted expressions as aggregation guidance may introduce undesirable inductive bias. 
To address the issues, we present a visual-linguistic contrastive loss that explicitly focuses on the alignment between video-level features and textual information. 
The loss accompanies two purposes: (1) Bridge the semantic gap between textual expressions and the corresponding object queries in the multi-modal joint embedding space. (2) Mitigate the bias caused by unrestricted textual features and emphasize the referred object representations.
As shown in \cref{fig:framework}, we transform the last stage of fused textual features $\mathcal{F}^{ef}_4$ from the vision-to-language stream (see \cref{eq:3}) to generate the textual guidance embedding $\mathcal{F}_{gud}$:
\begin{equation}
\mathcal{F}_{gud} = \mathrm{AveragePooling}\left( \mathcal{F}^{ef}_4\right), \mathcal{F}_{gud} \in \mathbb{R}^{D}.
\end{equation}
Then, we measure the similarity between the textual guidance embedding and video-level object queries by scaled dot product similarity: $\hat{y}_{sim} = \frac{\mathcal{O}^{v}\mathcal{F}_{gud}^\top}{\sqrt{D}}$.
To suppress undesired region response and highlight target object, we take softmax operation along the object query axis.
Finally, we compute the  contrastive loss $\mathcal{L}_{con}$ by matrix multiplication of $\hat{y}_{sim}$ and $y_{\tau}$, where $y_{\tau}$ is annotated to $1$ for the best predicted trajectory and $0$ for others:
\begin{equation}
\mathcal{L}_{con} = -  \mathrm{LogSoftmax}\left(\hat{y}_{sim}\right) \cdot y_{\tau}.
\end{equation}

\subsection{Instance Segmentation and Loss}

\paragraph{Prediction Heads}
As depicted in \cref{fig:framework}, there are three lightweight heads built on top of the semantic integration module. 
The classification head is a concatenation of three fully connected layers. It directly predicts class probability $\hat{p} \in \mathbb{R}^{T \times N_q \times (K+1)}$ for frame-level object queries, where $K$ is the number of classes. Note that if $K=0$, the role of the classification head is to judge whether the object is referred by  the text description. 
The box head comprises three sequential linear layers, which is designed to transform the object queries to normalized bounding box information $\hat{b} \in \mathbb{R}^{T \times N_q \times 4}$, \textit{i.e.}, center coordinates, width, and height. 
Similar to ~\cite{referformer}, we adopt dynamic convolution to output segmentation masks for each frame. Specifically, the mask head produces weights of $N_q$ dynamic kernels $\Omega = \{\omega_n\}_{n=1}^{N_q}$. Meanwhile, 
the encoded multi-scale cross-modal features $\{\mathcal{F'}^{vf}_i\}$ are stacked with the $4\times $ features from the visual backbone to form hierarchical decoding features.
An FPN structure takes the hierarchical features as input and outputs high-resolution semantic-aware features $\mathcal{F}_{seg} \in \mathbb{R}^{T\times \frac{H_0}{4} \times \frac{W_0}{4} \times D}$.
Finally, the dynamic kernels process the features $\mathcal{F}_{seg}$ based on the position information output from the box head and obtain the segmentation mask $\hat{m} \in \mathbb{R}^{T \times N_q \times \frac{H_0}{4} \times \frac{W_0}{4}}$, which can be formulated as:
\begin{equation}
\hat{m}_n=\left\{\hat{f}_n^{seg} \circledast \omega_i\right\}_{n=1}^{N_q}.
\end{equation}

\vspace{-5pt}
\paragraph{Instance Matching}
As described above, the prediction heads output the prediction trajectories of $N_q$ objects, denoted as $\hat{y} = \{\hat{y}_n\}$ and the n-th object trajectory prediction is:
\begin{equation}
    \hat{y}_n = \{\hat{p}_n^t, \hat{b}_n^t, \hat{m}_n^t\}_{t}^{T}.
\end{equation}
There is only one referred object in a video clip corresponding to the text descriptions. Therefore, we denote the ground truth object sequence as $y = \{p^t, b^t, m^t\}_{t}^T$ and search the best prediction trajectory $\hat{y}_{\sigma}$ via Hungarian algorithm~\cite{hungarian}. 

\vspace{-5pt}
\paragraph{Total Loss}
We supervise the trajectory prediction $\hat{y}_{\sigma}$ by four types of losses: (1) $\mathcal{L}_{mask}\left(y, \hat{y}_{\sigma}\right)$, the mask loss is a combination of Dice loss and binary focal loss, which is computed across frames. (2) $\mathcal{L}_{box}\left(y, \hat{y}_{\sigma}\right)$: the box loss aggregates the L1 loss and GIoU loss per-frame. (3) $\mathcal{L}_{cls}\left(y, \hat{y}_{\sigma}\right)$: the class loss is focal loss and  supervises the predicted object category. (4) $\mathcal{L}_{con}\left(y_{\tau}, \hat{y}_{sim} \right)$: visual-linguistic contrastive loss.
The total loss can be formulated as:
\begin{equation}
    \mathcal{L}_{total} = \lambda_{mask}\mathcal{L}_{mask} + \lambda_{box}\mathcal{L}_{box} + \lambda_{cls}\mathcal{L}_{cls} +
    \lambda_{con}\mathcal{L}_{con},
\end{equation}
where $\lambda$ is the scale factor to balance each loss.

\section{Experiment}
\subsection{Datasets and Metrics}
\textbf{Datasets.} We evaluate our model on four prevalent RVOS benchmarks: Ref-YouTube-VOS~\cite{urvos}, Ref-DAVIS17~\cite{davis}, A2D-Sentences, and JHMDB-Sentences~\cite{GavrilyukGLS18}. For detailed descriptions of the datasets please see the \cref{app:a}.

\vspace{-5pt}
\textbf{Metrics.}
Following ~\cite{mttr,referformer}, we measure the effectiveness of our model by criteria of Precision@K, Overall IoU, MeanIoU and MAP over 0.50:0.05:0.95 for A2D-Sentences and JHMDB-Sentences. Meanwhile, we adopt standard evaluation metrics: region similarity($\mathcal{J}$), contour accuracy ($\mathcal{F}$) and their average value ($\mathcal{J} \& \mathcal{F}$) on Ref-YouTube-VOS and Ref-DAVIS17. 

\vspace{-5pt}
\subsection{Implementation Details}
We take the pretrained Video Swin Transformer~\cite{videoswin} and RoBERTa~\cite{roberta}  as our encoder in default. Both the frame aggregation and object cluster parts of SIM consist of  three encoder and decoder layers. The number of frame-level queries $\mathcal{O}^{f}$ and video-level queries $\mathcal{O}^{v}$ are set as 20 in default. 
We feed the model windows of $w=8$ frames during training. 
The models are trained with eight 32GB V100 GPUs in default. The coefficients for losses are set as $\mathcal{\lambda}_{cls} = 2$, $\mathcal{\lambda}_{L1} = 2$, $\mathcal{\lambda}_{giou} = 2$, $\mathcal{\lambda}_{dice} = 2$, $\mathcal{\lambda}_{focal} = 5$, $\mathcal{\lambda}_{con} = 1$. 
Due to space limitations, please see the \cref{app:b} for more training details.

\vspace{-5pt}
\subsection{Main Results}

\begin{table*}[t]
    \centering
    \footnotesize
    \renewcommand\arraystretch{0.8}
    \setlength{\tabcolsep}{10pt}
    \begin{tabular}{l|c|ccc|ccc}
    \toprule
    \multirow{2}{*}{Method} & \multirow{2}{*}{Backbone} & \multicolumn{3}{c}{Ref-YouTube-VOS}  & \multicolumn{3}{c}{Ref-DAVIS17} \\ \cline{3-8} 
                            &                           & $\mathcal{J} \& \mathcal{F}$ & $\mathcal{J}$ & \multicolumn{1}{c|}{$\mathcal{F}$} & $\mathcal{J} \& \mathcal{F}$       & $\mathcal{J}$        & $\mathcal{F}$  \\ 
    \midrule
    
    URVOS~\cite{urvos} & ResNet-50 & 47.2 & 45.3 & 49.2 & 51.5 & 47.3 & 56.0 \\
    LBDT-4~\cite{lbdt} & ResNet-50 & 49.4 & 48.2 & 50.6 & - & - & - \\
    MTTR   ~\cite{mttr} & Video-Swin-T & 55.3 & 54.0 & 56.6 & - & - & - \\
    ReferFormer ~\cite{referformer} & Video-Swin-T   &56.0 & 54.8 & 57.3 & - & - & - \\   
    \rowcolor{gray!10} SOC (Ours) & Video-Swin-T   &\textbf{59.2} &\textbf{57.8} & \textbf{60.5} & \textbf{59.0} & \textbf{55.4} & \textbf{62.6} \\    
    \midrule
        \multicolumn{8}{c}{\textit{With Image Pretrain}} \\
        \midrule
    ReferFormer ~\cite{referformer} & Video-Swin-T & 59.4 & 58.0 & 60.9 & 59.7 & 56.6 & 62.8 \\
    ReferFormer ~\cite{referformer} & Video-Swin-B & 62.9 & 61.3 & 64.6 & 61.1 & 58.1 & 64.1 \\
    VLT~\cite{vlt} & Video-Swin-B & 63.8 & 61.9 & 65.6 & 61.6 & 58.9 & 64.3 \\
    \rowcolor{gray!10} SOC (Ours) & Video-Swin-T &\textbf{62.4} &\textbf{61.1} & \textbf{63.7} & \textbf{63.5} & \textbf{60.2} & \textbf{66.7} \\
    \rowcolor{gray!10} SOC (Ours) & Video-Swin-B & \textbf{66.0} & \textbf{64.1} & \textbf{67.9} & \textbf{64.2} & \textbf{61.0} & \textbf{67.4} \\
    
    \midrule
        \multicolumn{8}{c}{\textit{Joint Train}} \\
        \midrule
    ReferFormer & Video-Swin-T & 62.6 &59.9 &63.3 & - & - & - \\
    ReferFormer & Video-Swin-B & 64.9 &62.8 &67.0 & - & - & - \\
    \rowcolor{gray!10} SOC (Ours) & Video-Swin-T & \textbf{65.0} & \textbf{63.3} & \textbf{66.7} & \textbf{64.2} & \textbf{60.9} & \textbf{67.5} \\
    \rowcolor{gray!10} SOC (Ours) & Video-Swin-B & \textbf{67.3} & \textbf{65.3} & \textbf{69.3} & \textbf{65.8} & \textbf{62.5} & \textbf{69.1} \\
    \bottomrule
    
    \end{tabular}
    \caption{Comparison with the state-of-the-art methods on Ref-YouTube-VOS and Ref-DAVIS17 datasets. \textit{With Image Pretrain} denotes the models are first pretrained on RefCOCO~\cite{refcoco}, RefCOCO+~\cite{refcoco}, and RefCOCOg~\cite{grefcoco} datasets. \textit{Joint Train} indicates the models are trained with the combination of image datasets and video datasets.}
    \label{tab:ref}
    \vspace{-10pt}
\end{table*}

\textbf{Ref-YouTube-VOS \& Ref-DAVIS17.}
We compare our method to previous models on Ref-YouTube-VOS and Ref-DAVIS17 in \cref{tab:ref}.
With video-level multi-modal understanding, our SOC achieves new state-of-the-art performance among different training settings: train from scratch, with image pretrain, and joint train.
Without bells and whistles, our approach outperforms existing SOTA by about 3\% $\mathcal{J} \& \mathcal{F}$ under fair comparison. On Ref-DAVIS17, we directly report the results using the model trained on Ref-YouTube-VOS without finetune.


\begin{table*}[t]
\centering
\footnotesize
\renewcommand\arraystretch{0.8}
\setlength{\tabcolsep}{3.2pt}
\begin{tabular}{l | c | c c c c c |c c | c }

\toprule

\multirow{2}{*}{Method} & \multirow{2}{*}{Backbone} & \multicolumn{5}{c |}{Precision} & \multicolumn{2}{c |}{IoU} & \multirow{2}{*}{mAP} \\

 & & P@0.5 & P@0.6 & P@0.7 & P@0.8 & P@0.9  &Overall & Mean & \\

\midrule

Hu \textit{et al.}. ~\cite{hu16} & VGG-16 & 34.8 & 23.6 & 13.3 & 3.3 & 0.1 & 47.4 & 35.0 & 13.2   \\
Gavrilyuk \textit{et al}. ~\cite{GavrilyukGLS18}  & I3D & 47.5 & 34.7 & 21.1 & 8.0 & 0.2 & 53.6 & 42.1 & 19.8   \\ 
CMSA + CFSA ~\cite{cmfsa} & ResNet-101 & 48.7 & 43.1 & 35.8 & 23.1 & 5.2 & 61.8 & 43.2 & -   \\
ACAN ~\cite{wang2019acan} & I3D & 55.7 & 45.9 & 31.9 & 16.0 & 2.0  & 60.1 & 49.0 & 27.4  \\
CMPC-V ~\cite{liu2022cmpc} & I3D & 65.5 & 59.2 & 50.6 & 34.2 & 9.8  & 65.3 & 57.3 & 40.4 \\
ClawCraneNet ~\cite{liang2021clawcranenet} & ResNet-50/101 & 70.4 & 67.7 & 61.7 & 48.9 & 17.1 & 63.1 & 59.9 & -  \\
MTTR ~\cite{mttr} & Video-Swin-T & 75.4 & 71.2 & 63.8 & 48.5 & 16.9  & 72.0 & 64.0 & 46.1 \\
ReferFormer  ~\cite{referformer} & Video-Swin-T & 76.0 & 72.2 & 65.4 & 49.8 & 17.9  & 72.3 & 64.1 & 48.6  \\
\rowcolor{gray!10} SOC (Ours) & Video-Swin-T & \textbf{79.0} & \textbf{75.6} & \textbf{68.7} & \textbf{53.5} & \textbf{19.5}  & \textbf{74.7} & \textbf{66.9} & \textbf{50.4}  \\

\midrule
    \multicolumn{10}{c}{\textit{With Image Pretrain}} \\
    \midrule

ReferFormer~\cite{referformer} & Video-Swin-T & 82.8 & 79.2 & 72.3 & 55.3 & 19.3  & 77.6 & 69.6  & 52.8 \\

ReferFormer~\cite{referformer} & Video-Swin-B & 83.1 & 80.4 & 74.1 & 57.9 & 21.2  & 78.6& 70.3 & 55.0 \\

\rowcolor{gray!10} SOC (Ours) & Video-Swin-T & \textbf{83.1} & \textbf{80.6} & \textbf{73.9} & \textbf{57.7} & \textbf{21.8}  & \textbf{78.3} & \textbf{70.6} & \textbf{54.8} \\
\rowcolor{gray!10} SOC (Ours) & Video-Swin-B & \textbf{85.1} & \textbf{82.7} & \textbf{76.5} & \textbf{60.7} & \textbf{25.2} &  \textbf{80.7} & \textbf{72.5} &\textbf{57.3} \\

\bottomrule

\end{tabular}
\caption{Comparison with the state-of-the-art methods on A2D-Sentences.}
\label{lab:a2d}
\end{table*}

\textbf{A2D-Sentences \& JHMDB-Sentences.}
As shown in \cref{lab:a2d}, compared with the existing SOTA method ReferFormer~\cite{referformer}, our SOC achieves 50.4\% mAP and 66.9\% mIoU with the model trained from scratch, which gains a clear improvement of 1.8\% mAP and 2.8\% mIoU respectively. Furthermore, with image data pretraining, our method achieves new state-of-the-art on all metrics, \textit{e.g.}, +4.0\% P@0.9, +2.2\% mIoU, and +2.3\% mAP.  Due to the similarity of the datasets and space limitations, please see the \cref{app:jhmdb} for the comparison on JHMDB-Sentences~\cite{GavrilyukGLS18}. Our SOC also surpasses all existing methods on it.

\vspace{-5pt}
\subsection{Ablation Studies}
In this section, we conduct analysis experiments on the Ref-YouTube-VOS~\cite{urvos} benchmark using Video-Swin-Tiny as the visual backbone and train the model from scratch.

\vspace{-5pt}
\paragraph{Component Analysis}
We experiment with the effectiveness of the proposed
Video-level Object Cluster (VOC) and Visual-linguistic contrastive Learning (VL). As shown in \cref{tab:compont_ab}, both of them bring
performance improvement and their combination works better (+ 3.6\% $\mathcal{J} \& \mathcal{F}$), which shows the significance of video-level multi-modal understanding.

\begin{table*}[t]
\begin{minipage}[c]{\textwidth}
    \begin{minipage}{0.47\textwidth}
        \makeatletter\def\@captype{table}
        \centering
        \footnotesize
        \renewcommand\arraystretch{0.8}
        \setlength{\tabcolsep}{7pt}
        \begin{tabular}{l c c c c}
        \toprule
        VOC & VL & $\mathcal{J} \& \mathcal{F}$ &$\mathcal{J}$ & $\mathcal{F}$ \\
        \midrule
        & & 55.6 & 54.3 & 56.9 \\
        \checkmark &   &  58.0 & 56.5 & 59.6 \\
        &\checkmark   &  57.1 & 55.8 & 58.4 \\
        \checkmark &\checkmark   & \textbf{59.2} &\textbf{57.8} &\textbf{60.5} \\
        \bottomrule
        \end{tabular}
        \caption{Effectiveness of our proposed modules.}
        \vspace{10pt}
        \label{tab:compont_ab}
    \end{minipage}
    \hspace{0.01\textwidth}
    \begin{minipage}[c]{0.47\textwidth}
        \makeatletter\def\@captype{table}
        \centering
        \footnotesize
        \renewcommand\arraystretch{0.8}
        \setlength{\tabcolsep}{6pt}
        \begin{tabular}{l c c c}
        \toprule
        Fusion Strategy & $\mathcal{J} \& \mathcal{F}$ &$\mathcal{J}$ & $\mathcal{F}$ \\
        \midrule
        No Fusion   & 38.3 & 36.1  & 40.6 \\ 
        V2L   & 39.8 & 37.6  & 42.0 \\ 
        L2V   &  58.7 & 57.3 & 60.0 \\
        Both   &  \textbf{59.2} & \textbf{57.8} & \textbf{60.5} \\
        \bottomrule
        \end{tabular}
        \caption{Impact of different fusion strategies.}
        \vspace{10pt}
        \label{tab:fusion_ab}
    \end{minipage}
    \begin{minipage}[c]{0.31\textwidth}
        \makeatletter\def\@captype{table}  
        \centering
        \footnotesize
        \renewcommand\arraystretch{0.8}
        \setlength{\tabcolsep}{2.5pt}
        \begin{tabular}{l c c c}
        \toprule
         Query Num. & $\mathcal{J} \& \mathcal{F}$ &$\mathcal{J}$ & $\mathcal{F}$ \\
        \midrule
        10   & 58.0 & 56.6  & 59.5 \\ 
        15   &  58.1 & 56.7 & 59.6 \\
        20   &  \textbf{59.2} & \textbf{57.8} & \textbf{60.5} \\
        25   &  58.5 & 57.2 & 59.8 \\
        \bottomrule
        \end{tabular}
        \caption{Results of different object query numbers.}
        \label{tab:fusion_qn}
    \end{minipage}
    \hspace{0.01\textwidth}
    \begin{minipage}[c]{0.31\textwidth}
        \makeatletter\def\@captype{table}
        \centering
        \footnotesize
        \renewcommand\arraystretch{0.8}
        \setlength{\tabcolsep}{2.5pt}
        \begin{tabular}{l c c c}
        \toprule
        Frame Num.  & $\mathcal{J} \& \mathcal{F}$ &$\mathcal{J}$ & $\mathcal{F}$ \\
        \midrule
          3   &  57.9 & 56.5 & 59.3\\
          5   &  58.4 & 57.1 & 59.7 \\
          8   &  \textbf{59.2} & \textbf{57.8} & \textbf{60.5} \\
          10   &  59.1 & 57.8 & 60.3 \\
        \bottomrule
        \end{tabular}
        \caption{Results of different training frame numbers.}
        \label{tab:frame_num}
    \end{minipage}
    \hspace{0.01\textwidth}
    \begin{minipage}[c]{0.31\textwidth}
        \makeatletter\def\@captype{table}
        \centering
        \footnotesize
        \renewcommand\arraystretch{0.8}
        \setlength{\tabcolsep}{2.5pt}
        \begin{tabular}{l c c c}
        \toprule
        VOC Structure  & $\mathcal{J} \& \mathcal{F}$ &$\mathcal{J}$ & $\mathcal{F}$ \\
        \midrule
          None   &  57.1 & 55.8 & 58.4\\
          only encoder   &  58.3 & 56.9 & 59.8 \\
          only decoder   &  58.1 & 56.7 & 59.5 \\
          Both   &  \textbf{59.2} & \textbf{57.8} & \textbf{60.5} \\
        \bottomrule
        \end{tabular}
        \caption{The structure design of the VOC module.}
        \label{tab:voc_structure}
    \end{minipage}
\end{minipage}
\vspace{-10pt}
\end{table*}

\vspace{-5pt}
\paragraph{Fusion Strategy}
We investigate the effects of various fusion strategies. As illustrated in \cref{tab:fusion_ab}, the absence of Language to Vision fusion (L2V) yields significant deterioration. This is because the frame-level content aggregation without textual guidance would introduce background noise. If only utilizing the L2V fusion, the performance drops 0.5\% $\mathcal{J} \& \mathcal{F}$, demonstrating that visual content helps filter out irrelevant words from unconstrained descriptions.

\vspace{-5pt}
\paragraph{Segmentation Stability}
\begin{wrapfigure}{r}{5cm}
\vspace{-10pt}
        \centering
        \includegraphics[width=\linewidth,height=0.6\linewidth]{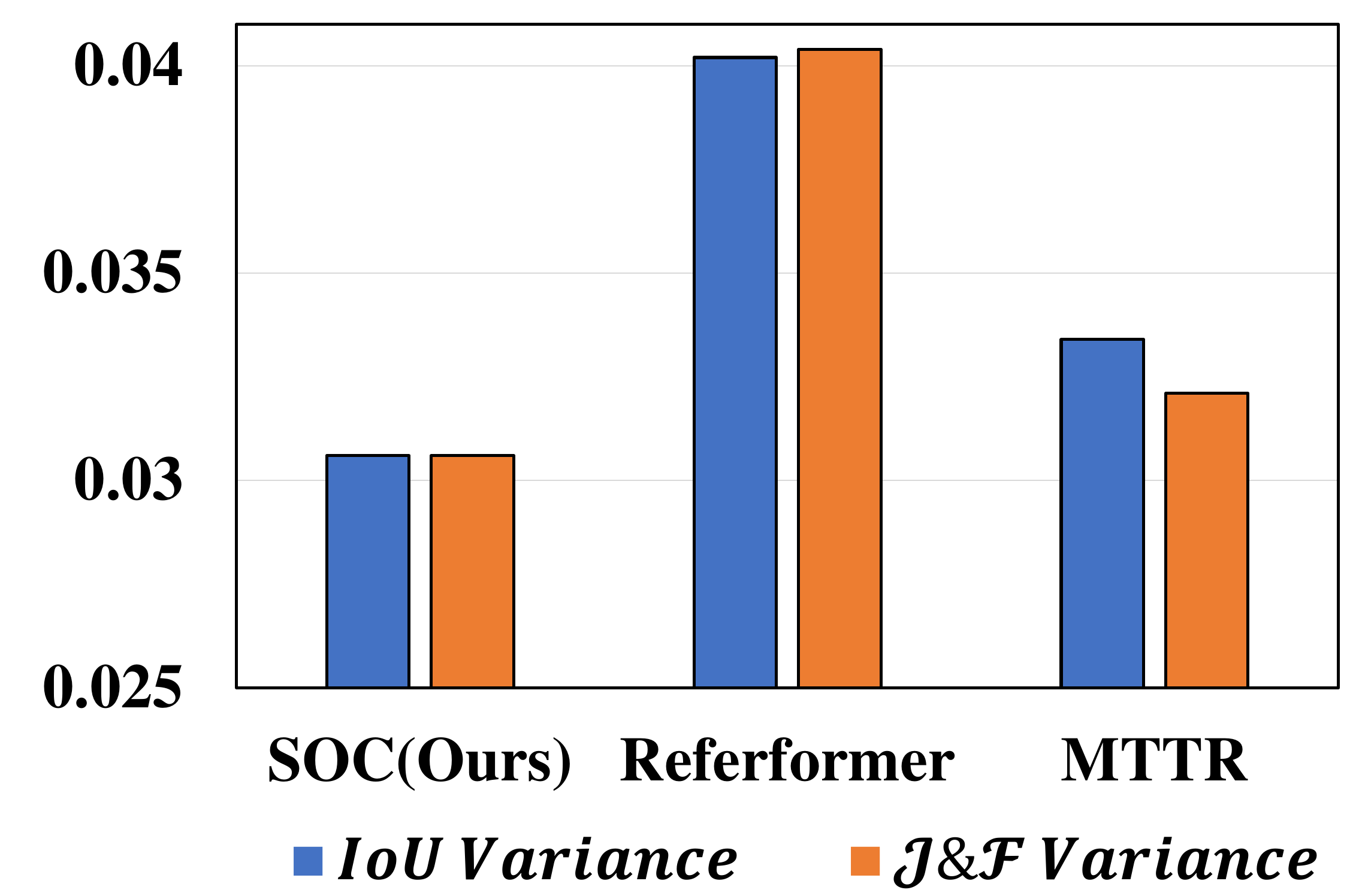}
    \vspace{-15pt}
    \caption{IoU and $\mathcal{J} \& \mathcal{F}$ variance.}
    \label{fig:variance}
    \vspace{-5pt}
\end{wrapfigure} 
Benefiting from video-level multi-modal understanding, our method yields more robust segmentation results compared to existing approaches. 
To quantify the temporal segmentation stability, we calculate the IoU and $\mathcal{J} \& \mathcal{F}$ variance between each frame in a video and count the average value on Ref-YouTube-VOS~\cite{urvos} dataset. Results are shown in \cref{fig:variance} (\textit{the lower is better}). It can be seen that with video object cluster module and visual-linguistic contrastive loss building the video-level multi-modal aligned space, our SOC can understand the overall video content better and output temporally stable segmentation results.
More qualitative analysis of temporal consistency is shown in the \cref{app:stability}.

\vspace{-5pt}
\paragraph{Object Query Number}
Although only one referred object is involved in a video, focusing on more potential regions is helpful due to the various video content. \cref{tab:fusion_qn} shows that an increased number of object queries allows the model to effectively group the relevant objects from the pool of candidate regions. The performance saturates when the query number is increased to a certain extent. We believe that excessive queries may cause misunderstandings when grouping objects across frames.          

\vspace{-5pt}
\paragraph{Frame Number}
\cref{tab:frame_num} shows results with different training clip frame numbers.
As expected, widening temporal context leads to performance gains yet causes expensive computation cost. For a fair comparison with existing methods, we take the frame number of 8 by default.

\vspace{-5pt}
\paragraph{Variants of VOC Structure}
Results in \cref{tab:voc_structure} illustrate the rationality of the structure of the VOC module. Without sufficient inter-frame interaction by the encoder or aggregation of target object information by the decoder, the segmentation performance will be degraded.

\vspace{-5pt}
\paragraph{Inference Speed}
In addition to superior segmentation performance, our method also achieves real-time inference. Specifically, our SOC runs at 32.3 FPS on single 3090 GPU, which significantly exceeds the existing SOTA method ReferFormer~\cite{referformer} (21.4 FPS).

\begin{figure*}
    \centering
    \includegraphics[width=\linewidth]{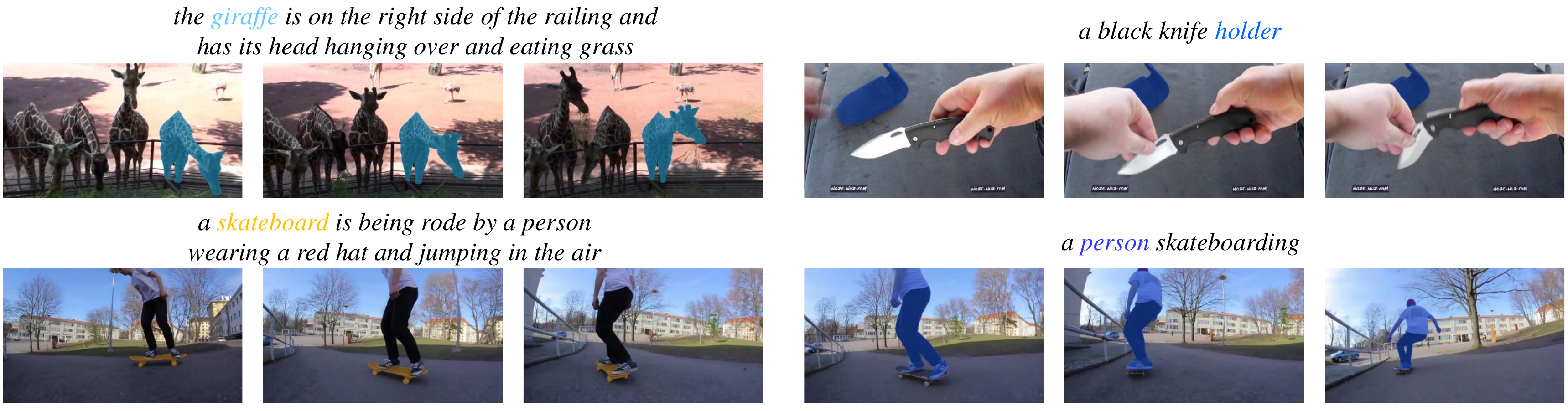}
    \vspace{-15pt}
    \caption{Segmentation results of our SOC. Best viewed in color.}
    \label{fig:case1}
\end{figure*}

\begin{figure*}
    \centering
    \includegraphics[width=\linewidth]{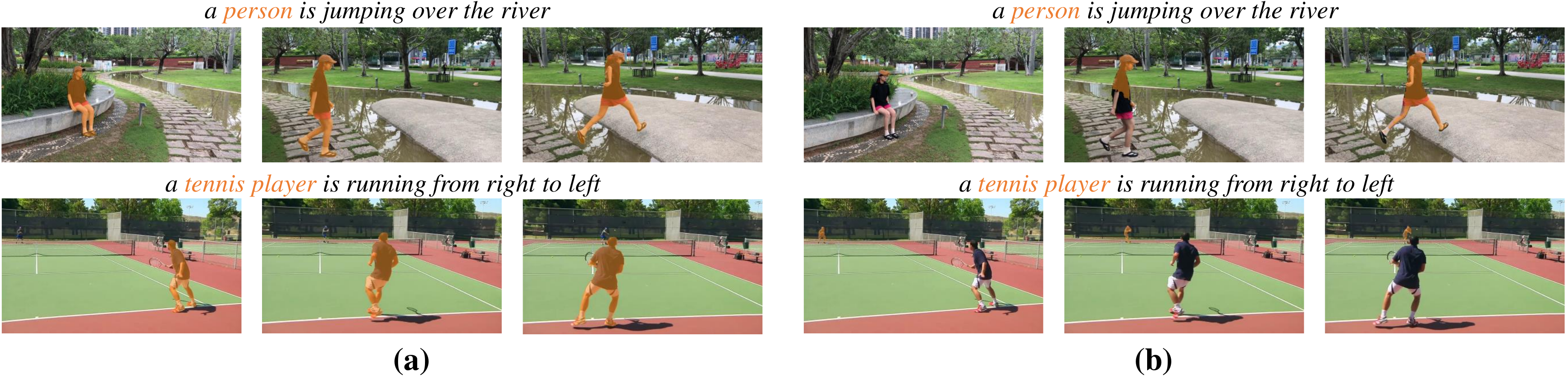}
    \vspace{-15pt}
    \caption{Visualization comparison using text expressions about temporal actions. (a) and (b) are segmentation results of our SOC and ReferFormer~\cite{referformer}, respectively.}
    \label{fig:case2}
    \vspace{-10pt}
\end{figure*}

\subsection{Qualitative Results}
\cref{fig:case1} demonstrates the effectiveness of SOC for complex scenarios segmentation, \textit{i.e.}, similar appearance, occlusion, and large variations. 
To further prove the \textit{adaptability} of our SOC for understanding text expressions about temporal action and variations, we design corresponding text descriptions that depict changed states of the object in temporal. 
As shown in \cref{fig:case2}, the global view enables SOC to understand such text  and segment the target object accurately. In contrast, ReferFormer~\cite{referformer} only recognizes the object in specific frames mentioned in the text descriptions and fails to comprehend the content.  More comparisons can be seen in the \cref{app:adaptability}.

\section{Conclusion}
In this paper, we propose a framework called SOC for RVOS to achieve video-level multi-modal understanding. By associating frame-level object embeddings with language tokens, we unify temporal modeling and cross-modal alignment into a simple architecture. Moreover, visual-linguistic contrastive learning is introduced to build the video-level multi-modal space. Extensive experiments show that our SOC remarkably outperforms existing state-of-the-art methods. 
Besides, video-level understanding also allows our SOC to handle text descriptions expressing temporal variations better.

\textbf{Limitations.} The proposed framework, SOC, has achieved video-level multi-modal understanding and excellent performance. However, there are some potential limitations, \textit{e.g.}, the current architecture cannot handle infinitely long videos. We think that devising explicit designs for long videos with complex scenarios is an interesting future direction.

{
\bibliographystyle{splncs04}
\bibliography{reference}
}


\newpage
\appendix

\section*{Appendix}
\section{Dataset Details} \label{app:a}
The A2D-Sentences dataset contains 3782 videos and each video has 3-5 annotated segmentation masks and JHMDB-Sentences totally comprises 928 videos, each of which is associated with a text description. For the large-scale dataset, Ref-YouTube-VOS has 3978 videos with about 15K text descriptions. The Ref-DAVIS17 contains 90 videos with 1,544 expressions, including 60 and 30 videos for training and validation respectively.

\section{Performance on JHMDB-Sentences} \label{app:jhmdb}
We also compare our SOC with existing methods on JHMDB-Sentences~\cite{GavrilyukGLS18} and the results are shown in \cref{tab:jhm}. Following ReferFormer~\cite{referformer}, we directly report the results utilizing the models trained on A2D-Sentences without finetune. It can be seen that our method achieves new state-of-the-art performance with different backbone and training settings. Compared to other benchmarks, the performance gains on this dataset are relatively small. This can be attributed to JHMDB's imprecise annotations generated by coarse human puppet model.

\begin{table*}[ht]
\centering
\footnotesize
\setlength{\tabcolsep}{2.5pt}
\begin{tabular}{l | c | c c c c c | c | c c}

\toprule

\multirow{2}{*}{Method} & \multirow{2}{*}{Backbone} & \multicolumn{5}{c |}{Precision} & \multirow{2}{*}{mAP} & \multicolumn{2}{c}{IoU}  \\

 & & P@0.5 & P@0.6 & P@0.7 & P@0.8 & P@0.9 & &Overall & Mean   \\

\midrule

Hu \textit{et al}. ~\cite{hu16} & VGG-16 & 63.3 & 35.0 & 8.5 & 0.2 & 0.0 & 17.8 & 54.6 & 52.8  \\
Gavrilyuk \textit{et al}. ~\cite{GavrilyukGLS18}  & I3D & 69.9 & 46.0 & 17.3 & 1.4 & 0.0 & 23.3 & 54.1 & 54.2  \\ 
CMSA + CFSA ~\cite{cmfsa} & ResNet-101 & 76.4 & 62.5 & 38.9 & 9.0 & 0.1 & - & 62.8 & 58.1 \\
ACAN ~\cite{wang2019acan} & I3D & 75.6 & 56.4 & 28.7 & 3.4 & 0.0 & 28.9 & 57.6 & 58.4  \\
CMPC-V ~\cite{liu2022cmpc} & I3D & 81.3 & 65.7 & 37.1 & 7.0 & 0.0 & 34.2 & 61.6 & 61.7  \\
ClawCraneNet ~\cite{liang2021clawcranenet} & ResNet-50/101 & 88.0 & 79.6 & 56.6 & 14.7 & 0.2 & - & 64.4 & 65.6  \\
MTTR  ~\cite{mttr} & Video-Swin-T & 93.9 & 85.2 & 61.6 & 16.6 & 0.1 & 39.2 & 70.1 & 69.8  \\
ReferFormer ~\cite{referformer} & Video-Swin-T & 93.3 & 84.2 & 61.4 & 16.4 & 0.3 & 39.1 & 70.0 & 69.3  \\

\rowcolor{gray!10} SOC(Ours) & Video-Swin-T & \textbf{94.7} & \textbf{86.4} & \textbf{62.7} & \textbf{17.9} & 0.1 & \textbf{39.7} &
\textbf{70.7} & \textbf{70.1}  \\

\midrule
    \multicolumn{10}{c}{\textit{With Image Pretrain}} \\
    \midrule

ReferFormer & Video-Swin-T & 95.8 & 89.3 & 66.8 & 18.9 & 0.2 & 42.2 & 71.9 & 71.0  \\

ReferFormer  & Video-Swin-B & 96.2 & 90.2 & 70.2 & 21.0 & 0.3 & 43.7 & 73.0 & 71.8  \\

\rowcolor{gray!10} SOC(Ours) & Video-Swin-T & \textbf{96.3} & \textbf{88.7} &\textbf{67.2} & \textbf{19.6} & 0.1 & \textbf{42.7} & \textbf{72.7} & \textbf{71.6} \\

\rowcolor{gray!10} SOC(Ours) & Video-Swin-B & \textbf{96.9} & \textbf{91.4} &\textbf{71.1} & \textbf{21.3} & 0.1 & \textbf{44.6} & \textbf{73.6} & \textbf{72.3} \\ 

\bottomrule
\end{tabular}
\caption{Comparison with the state-of-the-art methods on JHMDB-Sentences.}
\label{tab:jhm}
\end{table*}

\section{Comprehensive Evaluation}
\begin{wrapfigure}{r}{6cm}
\vspace{-10pt}
\centering
\includegraphics[width=\linewidth]{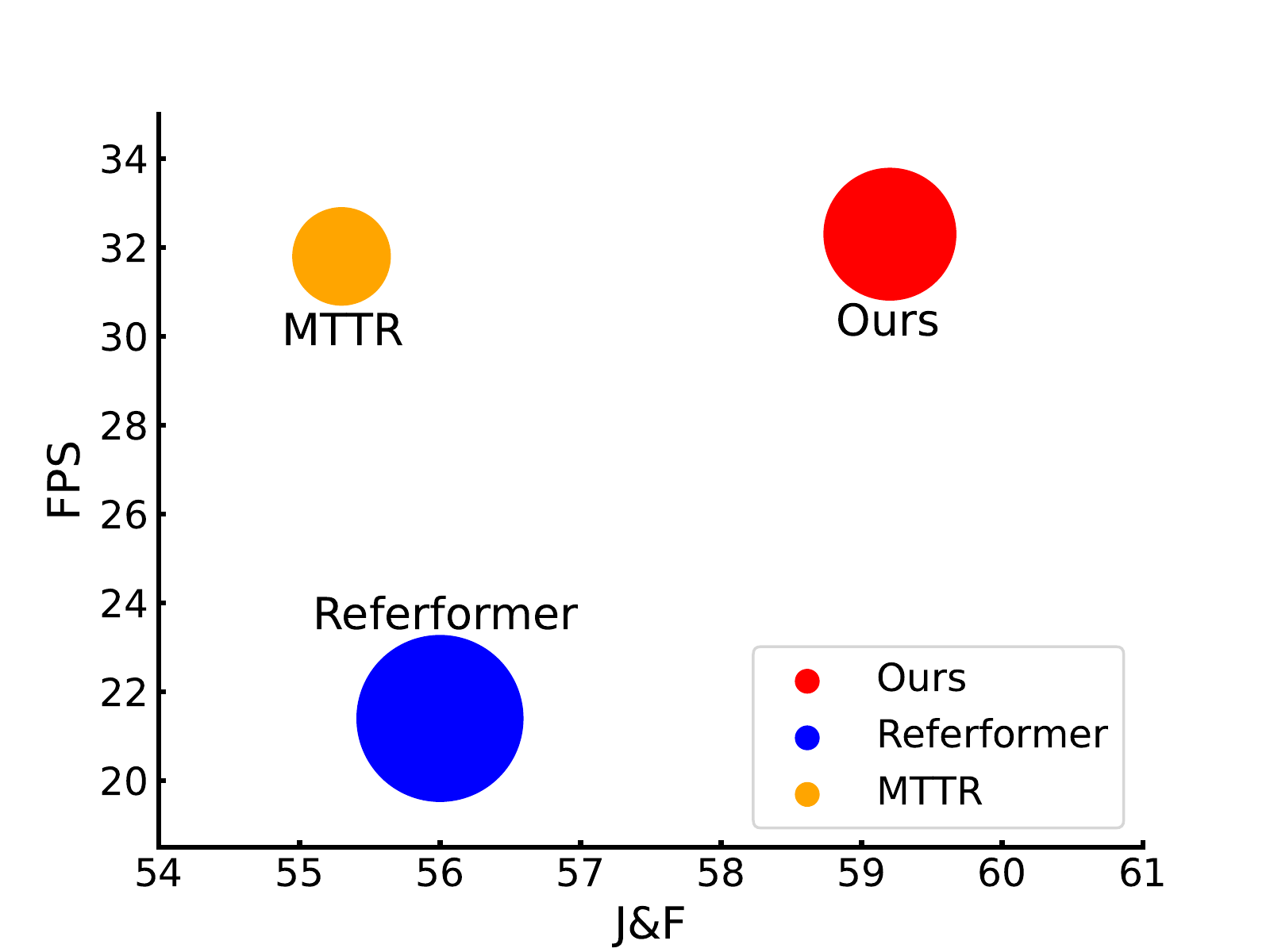}
    \vspace{-15pt}
    \caption{Performance \textit{vs} Inference Speed \textit{vs} Computation Cost.}
    \label{fig:flops}
\end{wrapfigure}
We comprehensively measure our method by different perspectives, \textit{e.g.}, performance, inference speed and computation cost under fair comparison. It is noted that the horizontal axis of \cref{fig:flops} denotes performance on Ref-YouTube-VOS, vertical axis is FPS and the radius of the circle represents the relative FLOPs. Compared with ReferFormer~\cite{referformer} (\textcolor{blue}{blue} \textcolor{blue}{$\bullet$}), Our method (\textcolor{red}{red} \textcolor{red}{$\bullet$}) achieves superior performance with faster inference speed and less computation cost. Although MTTR~\cite{mttr} (\textcolor{orange}{orange} \textcolor{orange}{$\bullet$})~\cite{mttr} has the lowest FLOPs, the lack of elaborate multi-modal fusion and temporal interaction significantly degrade the segmentation accuracy. In contrast, our method leverages video-level multi-modal understanding, which brings a significant increase in performance with only minimal computational costs.

\section{More Implementation Details} \label{app:b}

\paragraph{Training Settings}
Our models are trained with the AdamW optimizer using Pytorch. The weight decay is $1\times10^{-4}$.
The batch size is set to 56 for pretraining and 8 for main training.
The models are trained for 30 epochs. The initial learning rate is set to $1\times10^{-4}$ for Ref-YouTube-VOS and RefCOCO/+/g, $5\times10^{-5}$ for A2D-Sentences.
The learning rate decays by 10 for the backbone network.
During training, we apply RandomResize and Horizontal Flip for data augmentation.
Specifically, all frames are downsampled to 360$\times$640 for Ref-YouTube-VOS and RefCOCO/+/g, 320$\times$576 for A2D-Sentences.

\vspace{-5pt}
\paragraph{Inference Settings}
During inference, the input videos are downsampled to 320$\times$576 for A2D-Sentences dataset and 360p for other datasets. We directly output the segmentation masks without any post-process.

\section{Additional Qualitative Results}

\subsection{Query Visualization}
\begin{figure*}
    \centering
    \includegraphics[width=\linewidth]{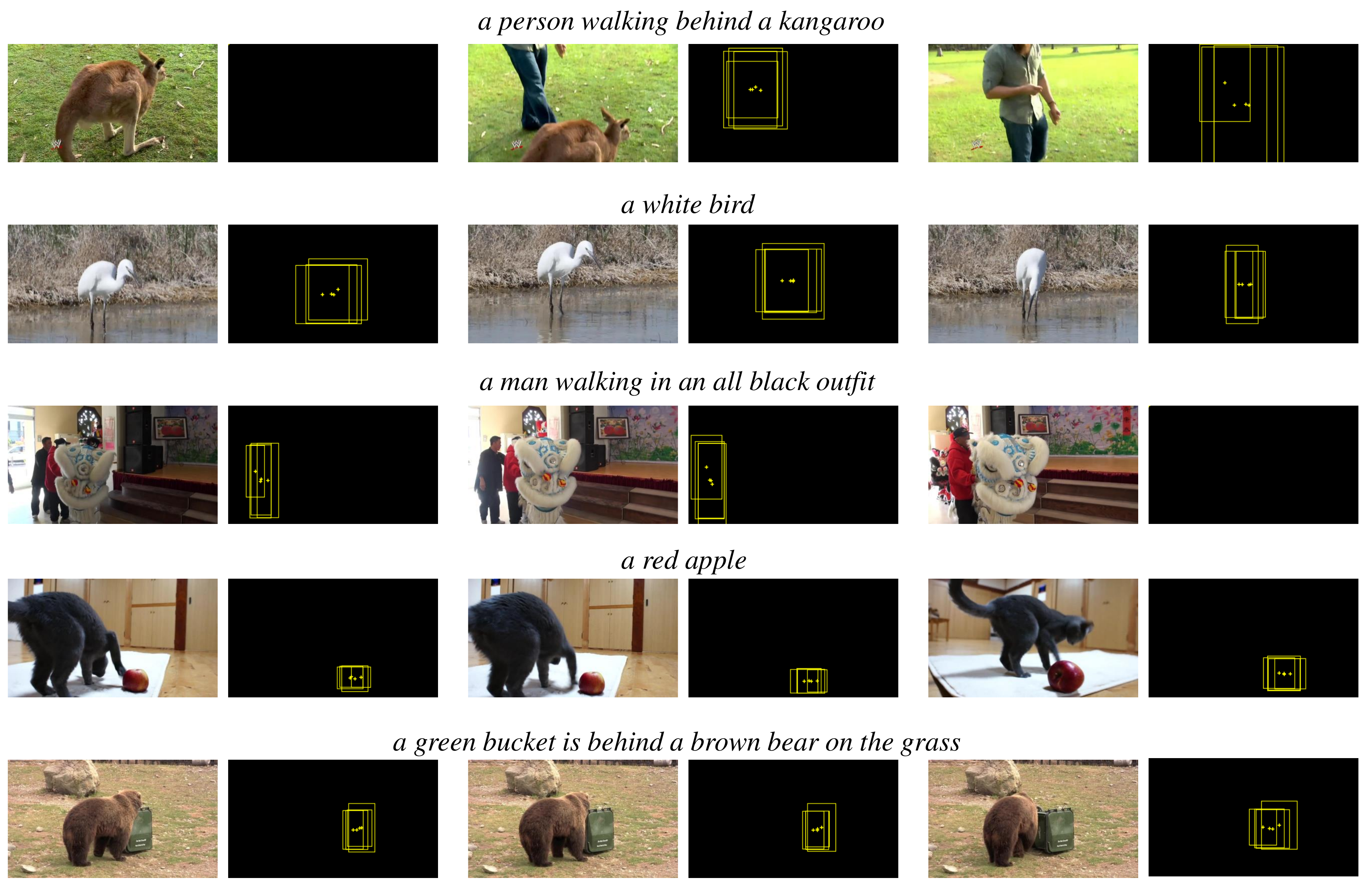}
    \vspace{-15pt}
    \caption{Visualization of the frame-level object query}
    \label{fig:query}
    \vspace{-10pt}
\end{figure*}

To demonstrate that the frame-level query embeddings can represent the referred object in a specific frame, we visualize the predicted bounding boxes corresponding to the query embeddings.
As illustrated in \cref{fig:query}, the majority of queries focus on regions of the referred object as expected. This indicates that the compact frame-level query embeddings indeed reflect object information and  subsequent video-level object cluster is performing temporal interaction for referred objects.

\subsection{Segmentation Stability Visualization}\label{app:stability}
\begin{figure*}
    \centering
    \includegraphics[width=\linewidth]{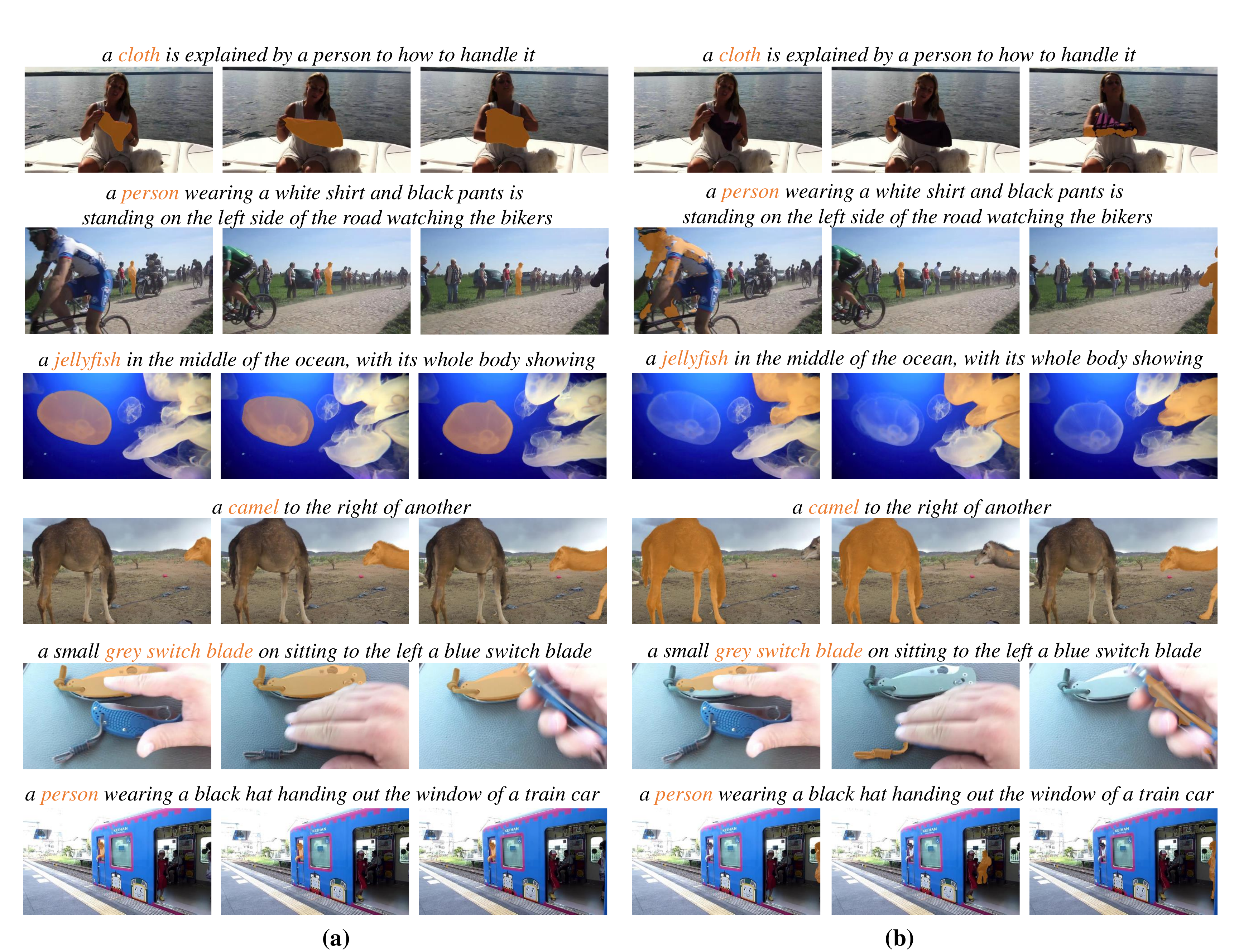}
    \vspace{-15pt}
    \caption{Visualization comparisons of segmentation stability between our SOC and existing state-of-the-art method ReferFormer~\cite{referformer}. (a) and (b) denote our SOC and ReferFormer, respectively.}
    \label{fig:stability}
    \vspace{-10pt}
\end{figure*}
The benchmark performance and IoU variance analysis in the main paper have proven the effectiveness and stability of our method.
Here we incorporate visual comparisons to further validate the segmentation stability of our model. In \cref{fig:stability} (a), benefiting from the global object view, SOC is capable of tracking the referred object across frames in coherence. On the contrary, ReferFormer~\cite{referformer}, the existing state-of-the-art method, may generate segmentation masks with high degree of variance, indicating that the frame-based paradigm fails to accurately understand the state of the object in the context of the entire video (see in \cref{fig:stability} (b)).     

\subsection{Adaptability for Texts Describing Temporal Variation} \label{app:adaptability}

\begin{figure*}
    \centering
    \includegraphics[width=\linewidth]{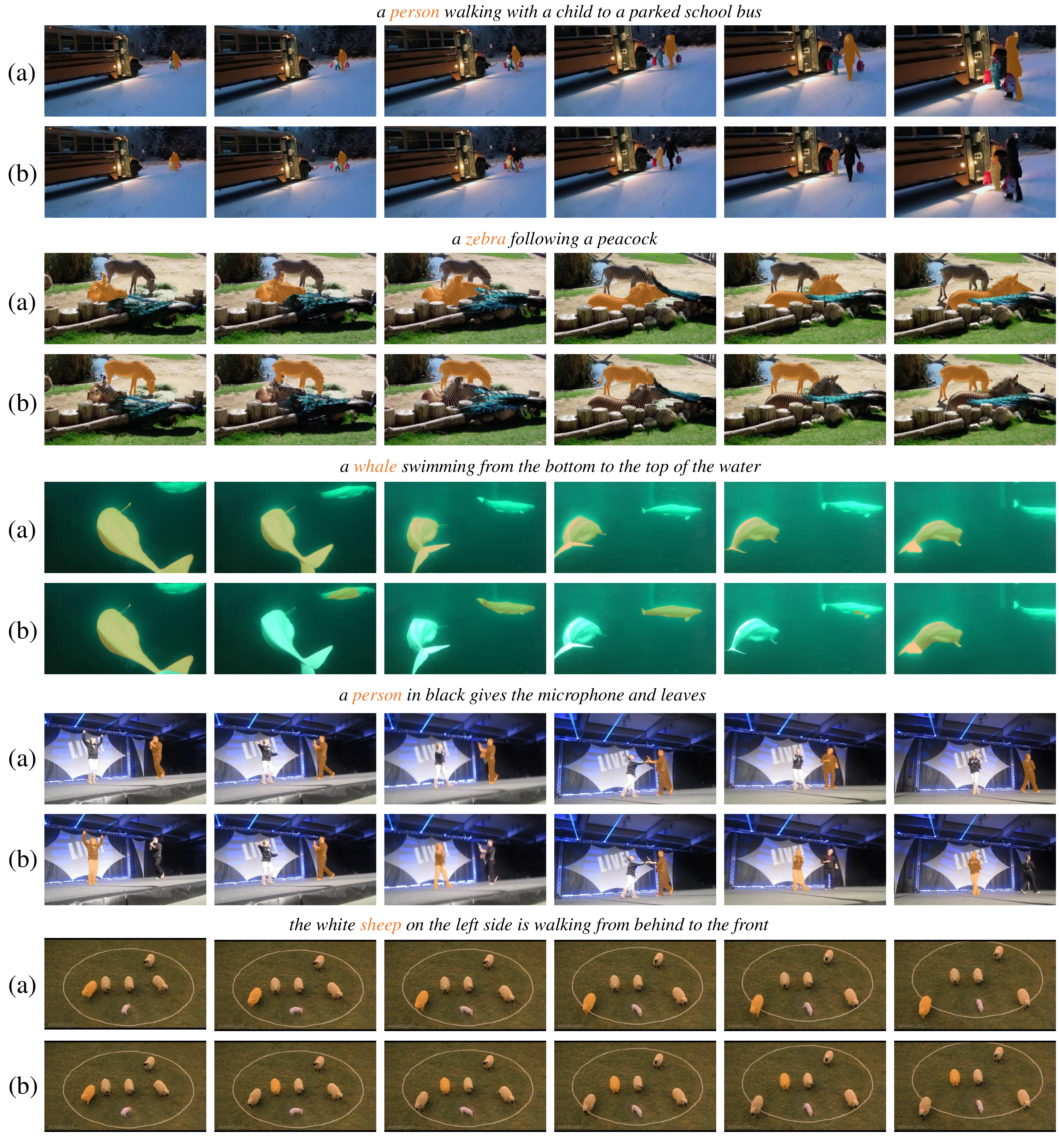}
    \caption{Visualization comparison using text expressions about temporal variation. (a) and (b) are segmentation results of our SOC and ReferFormer~\cite{referformer}, respectively.}
    \label{fig:temporal1}
    \vspace{-10pt}
\end{figure*}

\begin{figure*}[h]
    \centering
    \includegraphics[width=\linewidth]{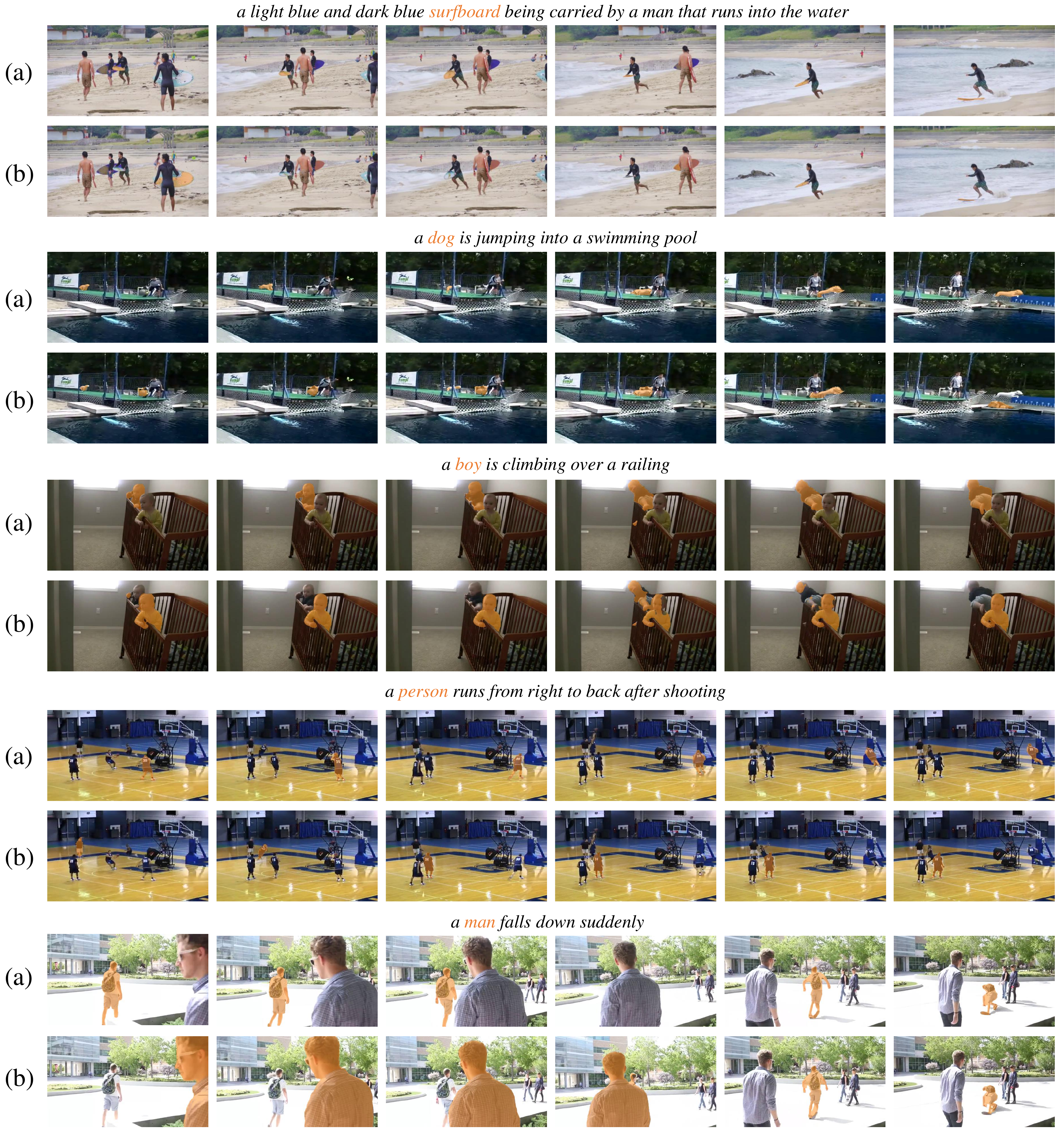}
    \caption{Visualization comparison using text expressions about temporal variation. (a) and (b) are segmentation results of our SOC and ReferFormer~\cite{referformer}, respectively.}
    \label{fig:temporal2}
\end{figure*}

\cref{fig:case2} in the main paper has shown some results to demonstrate that our SOC can better handle descriptions that focus on temporal variation.
Here we provide more cases to demonstrate the adaptability of our method to such text descriptions. \cref{fig:temporal1} and \cref{fig:temporal2} show the segmentation results of our SOC and ReferFormer, where (a) indicates the segmentation results by SOC and (b) represents the results by ReferFormer\cite{referformer}.

\section{Error Bar}
We have retrained our model several times on Ref-YouTube-VOS~\cite{urvos} dataset. The results demonstrate that the randomness of the model has little effect on the performance, \textit{i.e.}, the max deviation is about 0.5\% $\mathcal{J} \& \mathcal{F}$. 

\section{Broader Impact}
Malicious use of the RVOS model may lead to potential negative societal impacts, including but not limited to unauthorized surveillance or privacy-infringing tracking. However, we firmly believe that the task itself is neutral with positive implications, such as video editing and human-robot interaction.

\end{document}